\documentclass[twoside,leqno,twocolumn,dvipdfmx]{article}
\usepackage[letterpaper]{geometry}
\usepackage{ltexpprt}
\usepackage{bm}
\usepackage{amsmath}
\usepackage{amssymb}
\usepackage{dsfont}
\usepackage{algorithm}
\usepackage{algorithmic}
\usepackage{booktabs}
\usepackage{framed}
\usepackage[dvipdfmx]{graphicx}

\newenvironment{tsaligned}{%
  \begin{equation}\begin{aligned}%
}{%
  \end{aligned}\end{equation}%
}
\newenvironment{tsaligned*}{%
  \begin{equation*}\begin{aligned}%
}{%
  \end{aligned}\end{equation*}%
}

\newcommand{\tsincgra}[2]{{%
        \includegraphics[width=#1\textwidth]{#2}
}}
\newcommand{\tsja}[1]{}
\newcommand{\tsnote}[1]{}

  \newtheorem{definition}{Definition}
  {\endMakeFramed}

  {\endMakeFramed}

  {\endMakeFramed}

  \newenvironment{proposition-waku}
    {\begin{proposition}}
    {\end{proposition}}
  \newenvironment{definition-waku}
    {\begin{definition}}
    {\end{definition}}
  \newenvironment{lemma-waku}
    {\begin{lemma}}
    {\end{lemma}}
  \newenvironment{theorem-waku}
    {\begin{theorem}}
    {\end{theorem}}
  \newenvironment{corollary-waku}
    {\begin{corollary}}
    {\end{corollary}}

  \newcommand{\0}{{\bm{0}}}
  \newcommand{\1}{{\bm{1}}}

  \newcommand{\ve}{{\bm{e}}}

  \newcommand{\vh}{{\bm{h}}}
  
  \newcommand{\vk}{{\bm{k}}}

  \newcommand{\vq}{{\bm{q}}}

  \newcommand{\vs}{{\bm{s}}}

  \newcommand{\vt}{{\bm{t}}}
  \newcommand{\vu}{{\bm{u}}}

  \newcommand{\vw}{{\bm{w}}}
  
  \def\x{\bm{x}}

  \newcommand{\vA}{{\bm{A}}}

  \newcommand{\cB}{{\mathcal{B}}}

  \newcommand{\vI}{{\bm{I}}}

  \newcommand{\cK}{{\mathcal{K}}}

  \newcommand{\cL}{{\mathcal{L}}}

  \newcommand{\vM}{{\bm{M}}}
  \newcommand{\bN}{{\mathbb{N}}}

  \newcommand{\vO}{{\bm{O}}}
  
  \newcommand{\vQ}{{\bm{Q}}}

  \newcommand{\bR}{{\mathbb{R}}}

  \newcommand{\vU}{{\bm{U}}}

  \newcommand{\vW}{{\bm{W}}}

  \newcommand{\cY}{{\mathcal{Y}}}
  
  \newcommand{\cX}{{\mathcal{X}}}

  \newcommand{\valph}{{\bm{\alpha}}}
  \newcommand{\vbeta}{{\bm{\beta}}}

  \newcommand{\veta}{{\bm{\eta}}}
  \newcommand{\vlam}{{\bm{\lambda}}}

  \newcommand{\vmu}{{\bm{\mu}}}

  \newcommand{\vpi}{{\bm{\pi}}}
  \newcommand{\vrho}{{\bm{\rho}}}
  
  \newcommand{\vpsi}{{\bm{\psi}}}

  \newcommand{\vtau}{{\bm{\tau}}}

  \newcommand{\vLam}{{\bm{\Lambda}}}
  
  \newcommand{\vPsi}{{\bm{\Psi}}}

  \newcommand{\argmax}{\mathop{\textrm{argmax}}\limits}
  \newcommand{\argmin}{\mathop{\textrm{argmin}}\limits}

\newcommand{\qed}{\begin{flushright}q.e.d.\end{flushright}}

\makeatletter
\newcommand{\customlabel}[2]{%
\protected@write \@auxout {}{\string \newlabel {#1}{{#2}{}}}}
\makeatother

\begin{document}

\title{\Large
  Frank-Wolfe algorithm for learning SVM-type multi-category classifiers}
\author{Kenya Tajima\thanks{Gunma University} \and Yoshihiro Hirohashi\thanks{Individual} \and Esmeraldo Ronnie Rey Zara\footnotemark[1] \and Tsuyoshi Kato\footnotemark[1]}

\date{}

\maketitle


\fancyfoot[R]{\scriptsize{arXiv version.}}





\begin{abstract}
Multi-category support vector machine (MC-SVM) is one of the most popular machine learning algorithms. There are lots of variants of MC-SVM, although different optimization algorithms were developed for different learning machines. In this study, we developed a new optimization algorithm that can be applied to many of MC-SVM variants. The algorithm is based on the Frank-Wolfe framework that requires two subproblems, direction finding and line search, in each iteration. The contribution of this study is the discovery that both subproblems have a closed form solution if the Frank-Wolfe framework is applied to the dual problem. Additionally, the closed form solutions on both for the direction finding and for the line search exist even for the Moreau envelopes of the loss functions. We use several large datasets to demonstrate that the proposed optimization algorithm converges rapidly and thereby improves the pattern recognition performance. 
\end{abstract}
\section{Background and Motivation}

Multi-category classification is a task to assign an input object to one of pre-defined categories. Many supervised learning problems are reduced to the multi-category classification, although in the field of pattern recognition, the focus of many researches and theoretical analyses have been a simpler task, the binary classification, yielding the most successful machine learning algorithm, the support vector machine (SVM). In the 90's, the so-called one-versus-rest approach was employed to apply SVM to multi-category classification tasks. In the one-versus-rest approach, the learning task is divided into many independent optimization problems, and SVM is applied to each of the optimization problems. A drawback of the one-versus-rest approach is the inability for learning correlation among the categories. Crammer and Singer~\cite{Crammer-jmlr02} proposed an alternative method, which formulates the learning problem with a single optimization problem. This method is called the multi-category SVM (MC-SVM). Since the emergence of Crammer and Singer's MC-SVM, many variants such as the structured SVM~\cite{TsoJoaHofAlt05a}, SVM${}_{\textnormal{multi}}$~\cite{Joachims-icml2005}, top-k SVM~\cite{lapin-nips2015} have been developed. Structured SVM expanded the applicability of machine learning to a wide range including natural language parsing~\cite{Hare-pami2016} and the deformable part model for image analysis~\cite{Felzenszwalb-pami2010}, and the biological sequence alignment~\cite{TsoJoaHofAlt05a}. SVM${}_{\textnormal{multi}}$ provides a framework that directly learns the performance measures such as F1-score and precision/recall breakeven point, precision at k, and ROC score~\cite{Joachims-icml2005}. Top-k SVM is trained by minimizing the empirical risk based on top-$k$ error~\cite{lapin-nips2015}.

Learning machine cannot be practical without efficient and stable optimization algorithm. The above mentioned MC-SVM's variants are learned with different optimization algorithms, each of which is specialized to the corresponding learning machine. For example, cutting plain methods~\cite{Joachims-ml2009} were developed for learning the structured SVM and SVM${}_{\textnormal{multi}}$. Optimization algorithms for learning top-$k$ SVM were proposed by two research groups~\cite{lapin-nips2015,DejunChu-tnn18}, and both algorithms were based on the stochastic dual coordinate ascent (SDCA) method~\cite{Shalev-Shwartz2013jmlr}. However, the algorithms were derived from an incorrect theory, making both the algorithms fail to attain an optimum~\cite{KatHir-acml19a}. Kato and Hirohashi~\cite{KatHir-acml19a} considered applying Frank-Wolfe method~\cite{Frank1956} to the dual problem of top-$k$ SVM. Frank-Wolfe method is an iterative framework for convex optimization over a polyhedron and each iteration consists of the direction finding step and the line search step. Sub-linear convergence to the optimum is guaranteed if both the two steps are performed exactly~\cite{jaggi13icml}. Kato and Hirohashi~\cite{KatHir-acml19a} found that both the direction finding step and the line search step can be given in a closed form, and the computational time is within $O(mn^{3})$.

One of main contributions of this study is the finding that both the direction finding step and the line search step of Frank-Wolfe method are expressed in a closed form not only for top-$k$ SVM but also a wide range of the MC-SVM variants. In this paper, a condition for expressing the two steps in a closed form is clarified. Compared to gradient methods that are often employed for machine learning, the proposed Frank-Wolfe algorithm possesses no hyper-parameter such as a step size often requiring a manual tuning for optimization and guarantees the accuracy for the resulting solution. Due to the discovery of this study, an optimization algorithm that does not require a step size and can be terminated with a pre-defined accuracy becomes available for learning a variety of MC-SVM variants.

In addition, we extended our analysis to the Moreau envelope~\cite{Bertsekas99} of the loss function. The Moreau envelope is a trick that is widely used in the machine learning field. Taking the Moreau envelope makes the loss functions smooth and thereby accelerates optimization in general~\cite{Rennie-ijcai05,Zhang-icml04,lapin-cvpr2016}. In this study, we found that each step of Frank-Wolfe method can be expressed in a closed form even when taking the Moreau envelope of the loss function.

\textbf{Notation: }
We shall use the notation $\pi(j\,;\,\vs)\in [m]$
which is the index of the $j$-th largest component
in a vector $\vs\in\bR^{m}$. 
When using this notation, the vector $\vs$ is omitted
if there is no danger of confusion. Namely,
for a vector $\vs\in\bR^{m}$, we can write 
\(
  s_{\pi(1)}\ge s_{\pi(2)}\ge \dots \ge s_{\pi(m)}. 
\)
Let us define
$\vpi(\vs) := \left[ \pi(1\,;\,\vs),\dots,\pi(m\,;\,\vs)\right]^\top$
and introduce a notation for a vector with permutated components as
\(
  \vs_{\vpi(\vs)}
  :=
  \left[ s_{\pi(1)},\dots,s_{\pi(m)}\right]^\top. 
\)

We use $\ve_{i}$ to denote a unit vector where $i$-th entry is one.
The $n$-dimensional vector all of whose entries are one is denoted by $\1_{n}$.
We use an operator $\lVert\cdot\rVert_{\text{F}}$ to denote the Frobenius norm. 

\section{Methods and Technical Solutions}
\subsection{MC-SVM and its Variants}
In this section, we review MC-SVM and its several variants. 
Let us denote the discrete output space by $\cY:=\{1,\dots,m\}$
where $m$ is the number of categories. 
In the scenario of multi-category classification,
prediction is assignment of an input $\x\in\cX$ to one of elements in $\cY$, 
where $\cX$ is the input space.
Feature vectors are extracted not only from an input but also from
a candidate of categories.
Let $\vpsi:\cX\times\cY\to\bR^{d}$ be the feature extractor.
A typical implementation of the feature extractor is
$\vpsi(\x,j) := \ve_{j}\otimes\x$,
where $\otimes$ is the operator for the Kronecker product
and $\ve_{j}$ is here an $m$-dimensional unit vector.
Using the model parameter $\vw\in\bR^{d}$,  
prediction score for the category $j$ is given by
the inner product between the feature vector and the parameter vector,
i.e. $\left<\vpsi(\x,j),\vw\right>$.
Prediction of an input $\x\in\cX$ is done by computing
the prediction score for each category $j$, say $\left<\vpsi(\x,j),\vw\right>$,
and finding the maximal score among $m$ prediction scores.
The corresponding category is the prediction result.

We use $n$ training examples
$(\x_{1},y_{1}),\dots,(\x_{n},y_{n})\in\cX\times\cY$,
to determine the value of the model parameter $\vw\in\bR^{d}$.
MC-SVM tries to find the minimizer of the regularized
empirical risk defined as
\begin{tsaligned}\label{eq:P-general-def}
  P(\vw)
  :=
  \frac{\lambda}{2}
  \lVert\vw\rVert^{2}
  +
  \frac{1}{n}
  \sum_{i=1}^{n}
  \Phi(\vPsi(\x_{i})^\top\vw\,;\,y_{i}). 
\end{tsaligned}
where $\vPsi(\x_{i})$ is the horizontal concatenation
of $m$ feature vectors
(i.e.~$\vPsi(\x):=\left[\vpsi(\x,1),\dots,\vpsi(\x,m)\right] \in\bR^{d\times m}$);
$\lambda$ is a positive constant called the \emph{regularized parameter}; 
$\Phi(\cdot\,;\,y):\bR^{m}\to\bR$ is a loss function.
For MC-SVM, the \emph{max hinge loss} $\Phi_{\text{mh}}(\cdot\,;\,y)$
is adopted for $\Phi(\cdot\,;\,y)$.
Using the Kronecker delta $\delta_{\cdot,\cdot}$, 
the max hinge loss is defined as
\begin{tsaligned}\label{eq:Phi-mh-def}
  \Phi_{\text{mh}}(\vs\,;\,y)
  :=
  \max_{j\in[m]}
  \left(
  s_{j}-s_{y}+1-\delta_{j,y}
  \right). 
\end{tsaligned}

\textbf{Fenchel dual: }
Function $D:\bR^{m\times n}\to\bR$ defined as
\begin{tsaligned}
  D(\vA)
  :=
  -\frac{\lambda}{2}
  \lVert\vw(\vA)\rVert^{2}
  -
  \frac{1}{n}
  \sum_{i=1}^{n}
  \Phi^{*}(-\valph_{i}\,;\,y_{i}). 
\end{tsaligned}
is a Fenchel dual to the regularized empirical risk $P:\bR^{d}\to\bR$,
where
$\vA:=\left[\valph_{1},\dots,\valph_{n}\right]\in\bR^{m\times n}$; 
$\vw(\vA):=\frac{1}{\lambda n}\sum_{i=1}^{n}\vPsi(\x_{i})\valph_{i}$; 
$\Phi^{*}(\cdot\,;\,y)$ is the \emph{convex conjugate} of $\Phi(\cdot\,;\,y)$. 
The optimal solution of the primal variable $\vw_{\star}$ is obtained by $\vw_{\star}:=\vw(\vA_{\star})$ where $\vA_{\star}$ is the maximizer of the dual objective $D(\vA)$. The gap $P(\vw(\vA))-D(\vA)$ is non-negative for any $\vA\in\bR^{m\times n}$ and vanishes at the optimum $\vA=\vA_{\star}$. 
From this fact, we can terminate the iterations for optimization when $P(\vw(\vA))-D(\vA)\le \epsilon$ with a pre-defined small positive constant $\epsilon$. Then, the primal error $P(\vw(\vA))-P(\vw_{\star})$ is guaranteed not to be over $\epsilon$.

\textbf{Structured SVM: }
In the structured SVM, a non-negative loss $\Delta_{\hat{y}}^{(i)}$ for the $i$-th training example is arbitrarily designed for the case that the $i$-th training example is predicted as $\hat{y}\in\cY$ (i.e.~$\argmax_{j\in[m]}\left<\vw_{j},\vpsi(\x_{i},j)\right>=\hat{y}$). This is contrastive to Crammer and Singer's MC-SVM that adopts the \emph{convex surrogate} of the 0/1 loss that always suffers a unit loss for a mistake.
Structured SVM employs the convex surrogate of $\Delta_{\hat{y}}^{(i)}$, defined as:
\begin{tsaligned}\label{eq:sh-def}
  \Phi_{i,\text{sh}}(\vs\,;\,y)
  :=
  \max_{j\in[m]}
  \left( s_{j}-s_{y}+\Delta_{j}^{(i)}\right). 
\end{tsaligned}

\textbf{Unweighted Top-$k$ SVM: }
The unweighted top-$k$ SVM is a variant of MC-SVM. While MC-SVM assumes that a single category is assigned to an input, the unweighted top-$k$ SVM assigns $k$ categories to an input. Prediction results are interpreted so that one of predicted $k$ categories will be the category of the input. Given an input $\x\in\cX$, the set of $k$ categories are chosen as $\{ \pi(1;\vs),\dots,\pi(k;\vs)\}$ where $\vs$ is the prediction score vector (i.e. $\vs=\left[s_{1},\dots,s_{m}\right]^\top := \vPsi(\x)^\top\vw$). 
For training such a classifier, the loss function is designed as
\begin{tsaligned}\label{eq:utk-def}
  \Phi_{utk}(\vs\,;\,y)
  :=
  \max\left\{0,\frac{1}{k}\sum_{j=1}^{k}
  (\vs-s_{y}\1 + \1 - \ve_{y})_{\pi(j)}. 
  \right\}
\end{tsaligned}
This is called the \emph{unweighted top-$k$ hinge loss}.

\textbf{Unweighted Usunier SVM: }
Similar to the unweighted top-$k$ SVM, the unweighted Usunier SVM trains the classifier that performs top-$k$ prediction, but the loss function is slightly different. The empirical risk for learning the unweighted Usunier SVM consists of the following loss function:
\begin{tsaligned}\label{eq:uu-def}
  \Phi_{uu}(\vs\,;\,y)
  :=
  \frac{1}{k}\sum_{j=1}^{k}
  \max\left\{0,
  (\vs-s_{y}\1 + \1 - \ve_{y})_{\pi(j;\vs-\ve_{y})}
  \right\}. 
\end{tsaligned}
This loss function is called the Usunier loss.
The original loss function developed by Usunier et al~\cite{Usunier2009} is
devised for ranking prediction. Lapin et al~\cite{lapin-nips2015} redesigned
their loss function for top-$k$ prediction.

\textbf{Weighted Top-$k$ SVM: }
Using a constant weight vector
$\vrho=\left[\rho_{1},\dots,\rho_{m}\right]^{\top}\in\bR^{m}$
such as $\rho_{1}\ge\dots\ge\rho_{m-1}\ge\rho_{m}=0$,
Kato and Hirohashi extended the unweighted top-$k$ hinge loss
to the weighted version:
\begin{tsaligned}\label{eq:wtk-def}
  \Phi_{\text{wtk}}(\vs\,;\,y)
  :=
  \max\left\{
  0,
  \sum_{j=1}^{m}
  \left( \1_{m} - \ve_{y} + \vs - s_{y}\1_{m}
  \right)_{\pi(j)}
  \rho_{j}
  \right\}. 
\end{tsaligned}
They called this function~\eqref{eq:wtk-def}
the \emph{weighted top-$k$ hinge loss}. 

\textbf{Weighted Usunier SVM: }
The weighted version of the Usunier loss can be considered.
The \emph{weighted Usunier loss} function is defined as
\begin{tsaligned}\label{eq:wu-def}
  \Phi_{\text{wu}}(\vs\,;\,y)
  :=
  \sum_{j=1}^{m}
  \max\left\{
  0,
  \left( \1_{m} - \ve_{y} + \vs - s_{y}\1_{m}
  \right)_{\pi(j)}
  \right\}
  \rho_{j}. 
\end{tsaligned}
where 
$\vrho=\left[\rho_{1},\dots,\rho_{m}\right]^{\top}\in\bR^{m}$
is a constant vector 
such as $\rho_{1}\ge\dots\ge\rho_{m-1}\ge\rho_{m}=0$. 
\subsection{Max Dot Over Simplex-Type Loss Functions}
In this section, learning machines targeted by our learning algorithm are formulated. The learning machine trains a classifier by minimizing the regularized empirical risk given in \eqref{eq:P-general-def}. In the learning algorithm presented in the next section, the loss function appearing in the expression of the regularized empirical risk is assumed to be the \emph{max dot over simplex-type} (\emph{mdos-type}) defined below.

\begin{definition}
  Function $\Phi:\bR^{m}\to\bR$ is said to be mdos-type if
  there exists a simplex $\cB$ such that 
  $\forall y \in[m]$, $\forall \vs\in\bR^{m}$,
  \begin{tsaligned}
    \Phi(\vs\,;\,y)
    =
    \max_{\vbeta\in\cB}
    \left<
    \vbeta,
    \1-\ve_{y}+\vs-s_{y}\1
    \right>. 
  \end{tsaligned}
\end{definition}

In the previous section, six loss functions, the max hinge loss, the structured hinge loss, the unweighted top-$k$ hinge loss, the unweighted Usunier loss, the weighted top-$k$ hinge loss, the weighted Usunier loss, were described. It can be shown that all these six loss functions are mdos-type. In what follows, the corresponding simplexes are presented. 

\begin{itemize}
\item The simplex for the max hinge loss and the structured hinge loss is 
  $\cB_{\text{sh}} := \Delta(1)$, where
  \begin{tsaligned}
    \Delta(r)
    :=
    \left\{
    \vbeta\in\bR_{+}^{m}
    \,\middle|\,
    \lVert\vbeta\rVert_{1} \le r 
    \right\}.
  \end{tsaligned}
\item The simplex for the unweighted top-$k$ loss is
  $\cB_{\text{utk}} := \Delta_{\text{tk}}(k,1)$, where
  \begin{tsaligned}
    \Delta_{\text{tk}}(k,r)
    :=
    \left\{
    \vbeta\in\Delta(r)
    \,\middle|\,
    \vbeta\le \frac{\lVert\vbeta\rVert_{1}}{k}\1
    \right\}
  \end{tsaligned}
\item The simplex for the unweighted Usunier loss~\eqref{eq:uu-def} is
  $\cB_{\text{uu}} := \Delta_{\text{u}}(k,1)$, where
  \begin{tsaligned}
    \Delta_{\text{u}}(k,r) :=
    \left\{
    \vbeta\in\Delta(r)    
    \,\middle|\,
    \vbeta\le \frac{1}{k}\1
    \right\}
  \end{tsaligned}
\item The simplex for the weighted top-$k$ hinge loss ~\eqref{eq:wtk-def} is 
\begin{tsaligned}\label{eq:wtksplx-def}
  &
  \cB_{\text{wtk}}
  :=
  \Big\{
  \vbeta\in\bR^{m}
  \,\Big|\,
  \exists \zeta\in\bR, \,\,
  \forall \ell\in[L],
  \\
  &
  \exists \vlam_{\ell}\in\Delta_{\text{tk}}(k_{\ell},\rho'_{\ell}k_{\ell}),
  \,
  \zeta = \frac{\left<\1,\vlam_{\ell}\right>}{k_{\ell}\rho'_{\ell}},
  \,
  \vbeta = \sum_{\ell=1}^{L}\vlam_{\ell} 
  \Big\}. 
\end{tsaligned}
  
\item The simplex of the weighted Usunier loss~\eqref{eq:wu-def} is
\begin{tsaligned}\label{eq:wusplx-def}
  \cB_{\text{wu}}
  :=
  &
  \Big\{
  \vbeta\in\bR^{m}_{+}
  \,\Big| \,
  \,
  \forall \ell\in[L],\,
  \\
  &
  \exists \vlam_{\ell} \in \Delta_{\text{uu}}(1/\rho'_{\ell},k_{\ell}\rho'_{\ell}),
  \,
  \vbeta \le \sum_{\ell=1}^{L}\vlam_{\ell} 
      \Big\}. 
\end{tsaligned}
\end{itemize}

Therein, the variables $L$ and $\rho'_{1},\dots,\rho'_{L}$ used
in \eqref{eq:wtksplx-def} and \eqref{eq:wusplx-def} are defined
as follows. The variable $L$ takes a natural number representing
the cardinality of the set
\begin{tsaligned}
  \cK := \left\{ k \in [m]\,\middle|\, \rho_{k}>\rho_{k+1} \right\}. 
\end{tsaligned}
Denote by $k_{1},\dots,k_{L}$ the entries in $\cK$ sorted as
$1 \le k_{1} < \dots < k_{L} < m$.
The rest of the variables
$\rho'_{1},\dots,\rho'_{L}$
are defined as
$\rho'_{\ell}:=\rho_{k_{\ell}}$ for $\ell=1,\dots,L$.
The above results are summarized in the following theorem.
\begin{theorem-waku}\label{thm:five-losses-mdos-typ}
  Either of four loss functions
  $\Phi_{\text{utk}}(\cdot;y)$,
  $\Phi_{\text{uu}}(\cdot;y)$,
  $\Phi_{\text{wtk}}(\cdot;y)$, and
  $\Phi_{\text{wu}}(\cdot;y)$
  is mdos-type.
  Loss function
  $\Phi_{i,\text{sh}}(\cdot\,;\,y)$ is mdos-type when
  $\Delta_{j}^{(i)}=1-\delta_{y,j}$
  where $\delta_{\cdot,\cdot}$ is the Kronecker delta. 
\end{theorem-waku}
The detail of the proof for Theorem~\ref{thm:five-losses-mdos-typ}
is given in Section~\ref{sec:proof-for-thm-mdos}. 
\subsection{Frank-Wolfe Algorithm}
\label{s:fw}
In this section, an optimization algorithm for learning MC-SVM is presented. Here, the loss function $\Phi$ appearing in the regularized empirical loss is supposed to be mdos-type. The optimization algorithm presented here is the Frank-Wolfe method maximizing the dual objective $D(\vA)$. Each iteration of the Frank-Wolfe method consists of the direction finding step and the line search step. Denote by
$\vA^{(t)}=\left[\valph_{1}^{(t)},\dots,\valph_{n}^{(t)}\right]$
the dual variable at $t$-th iteration. The direction finding step solves the following linear program: 
\begin{tsaligned}\label{eq:dirfind-step-fwgeneral}
  \vU^{(t-1)} \in \argmax_{\vU\in\text{dom}(-D)}
  \left< \nabla D(\vA^{(t-1)}), \vU \right>. 
\end{tsaligned}
At the line-search step, 
a solution maximizing $D(\vA)$ over the line segment
between two points, $\vA^{(t-1)}$ and $\vU^{(t-1)}$, is found: 
\begin{tsaligned}\label{eq:lnsrch-step-fwgeneral}
  \gamma^{(t-1)}
  :=
  \argmax_{\gamma\in[0,1]} D \left( (1-\gamma)\vA^{(t-1)}+\gamma\vU^{(t-1)}\right). 
\end{tsaligned}
Using the solutions to the two subproblems,
say $\vU^{(t-1)}$ and $\gamma^{(t-1)}$, the dual variable
is updated as 
\begin{tsaligned}
  \vA^{(t)}
  :=
  (1-\gamma^{(t-1)})\vA^{(t-1)}+\gamma^{(t-1)}\vU^{(t-1)}. 
\end{tsaligned}
So long as the two subproblems are solved exactly at each iteration, the sublinear convergence is guaranteed.
However, the algorithm would be impractical if each step could not be solved efficiently.

We first discuss how we can perform the direction finding step. 
Under the assumption that the loss function is mdos-type,
the effective domain $\text{dom}(-D)$ is a polyhedron. 
Hence, a general-purpose solver for linear programs can
be used for the direction finding step,
it takes a prohibitive computational cost
if resorting to a general-purpose solver at every iteration. 
In this study, we consider the following theorem.
\begin{theorem-waku}\label{thm:fw}
  Consider applying the Frank-Wolfe algorithm to the problem
  for maximizing $D(\vA)$ with respect to $\vA$.
  Assume that the loss function $\Phi(\cdot\,;\,y)$ to be mdos-type. 
  Then, both the direction finding step and the line search step
  are expressed in a closed form.
\end{theorem-waku}
See Section~\ref{sec:proof-for-thm-fw} for the proof of Theorem~\ref{thm:fw}. 
The concrete solution to the subproblem~\eqref{eq:dirfind-step-fwgeneral}
for the direction finding step is given as follows. 
The $i$-th column in the matrix $\vU^{(t-1)}$, say $\vu_{i}^{(t-1)}$,
is set to 
\begin{tsaligned}\label{eq:small-lp-sol-fwgeneral}
  \vu_{i}^{(t-1)}\in-\partial\Phi
  \left(
  \vPsi(\x_{i})^\top\vw(\vA^{(t-1)})
  \,;\,y_{i}
  \right)
\end{tsaligned}
where $\partial\Phi(\cdot\,;\,y)$ is the subdifferential of
$\Phi(\cdot\,;\,y)$.
An arbitrary subgradient can be taken even if
the set $\partial\Phi(\cdot\,;\,y)$ has multiple elements. 

The subproblem for the line search step \eqref{eq:lnsrch-step-fwgeneral}
is also solved in a closed-form solution as 
$\gamma^{(t-1)}=\max(0,\min(1,\hat{\gamma}^{(t-1)}))$
where
\begin{tsaligned}\label{eq:gamhat-fwgeneral}
  &\hat{\gamma}^{(t-1)}
  :=
  \frac{%
    \sum_{i=1}^{n}%
    \left<%
    \Delta\valph_{i}^{(t-1)}, 
    \ve_{y_{i}}-\vPsi(\x_{i})^\top\vw(\vA^{(t-1)})%
    \right>%
  }{%
    \lambda n\lVert\vw(\Delta\vA^{(t-1)})\rVert^{2}
  },
\end{tsaligned}
where 
$\Delta\valph_{i}^{(t-1)}$ is
the $i$-th column of the $m\times n$ matrix
$\Delta\vA^{(t-1)}:= \vU^{(t-1)}-\vA^{(t-1)}$.

\begin{figure*}[t!]
  \begin{center}
    {\footnotesize
  \begin{tabular}{lll}
    (a) Caltech101 & (b) CUB200 & (c) Flower102
    \\
    \tsincgra{0.3}{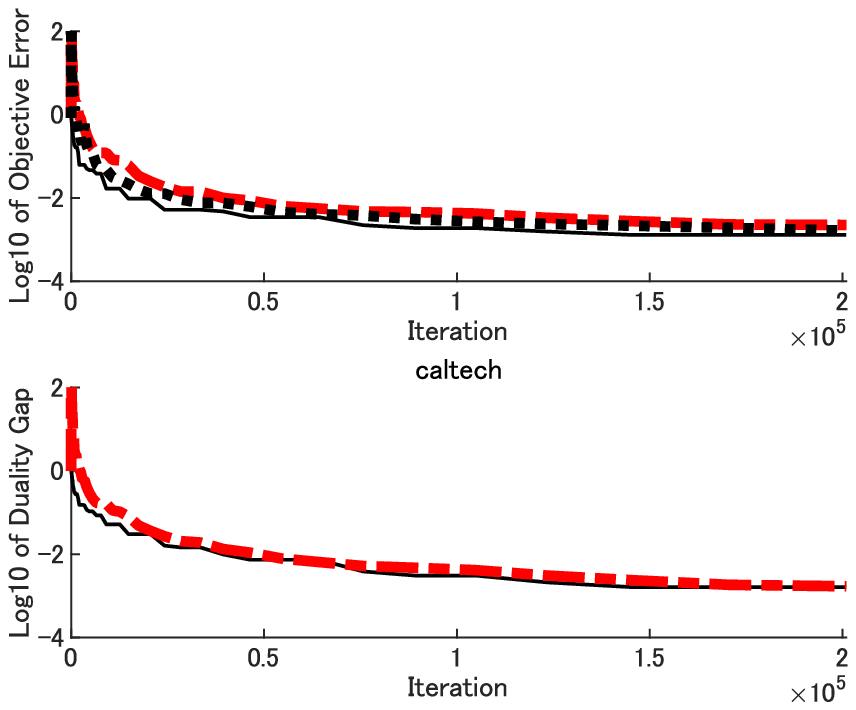}
    &
    \tsincgra{0.3}{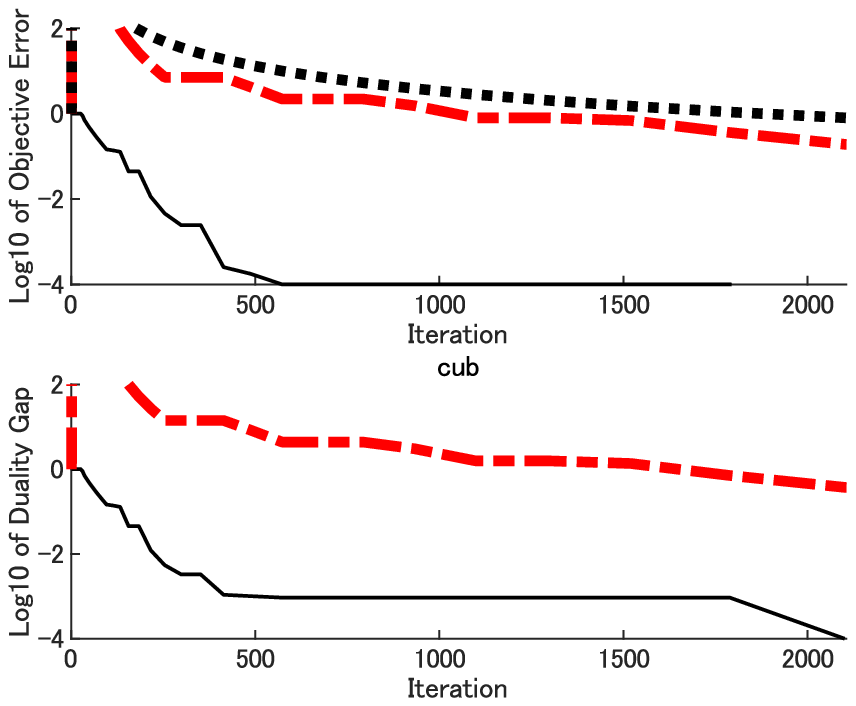}
    &
    \tsincgra{0.3}{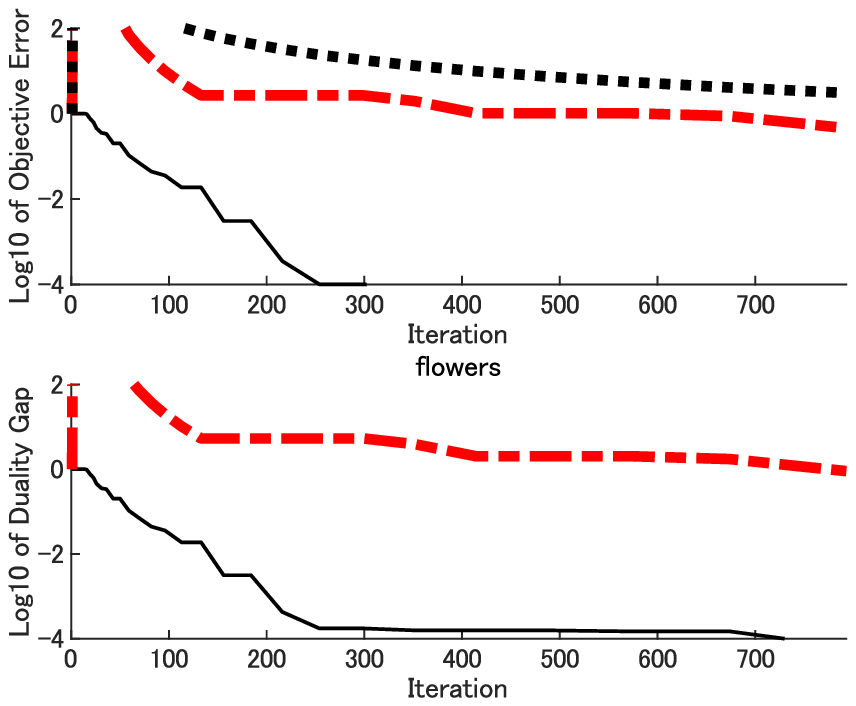}    
    \\
    \\
    (d) Indoor67 & (e) News20 & 
    \\
    \tsincgra{0.3}{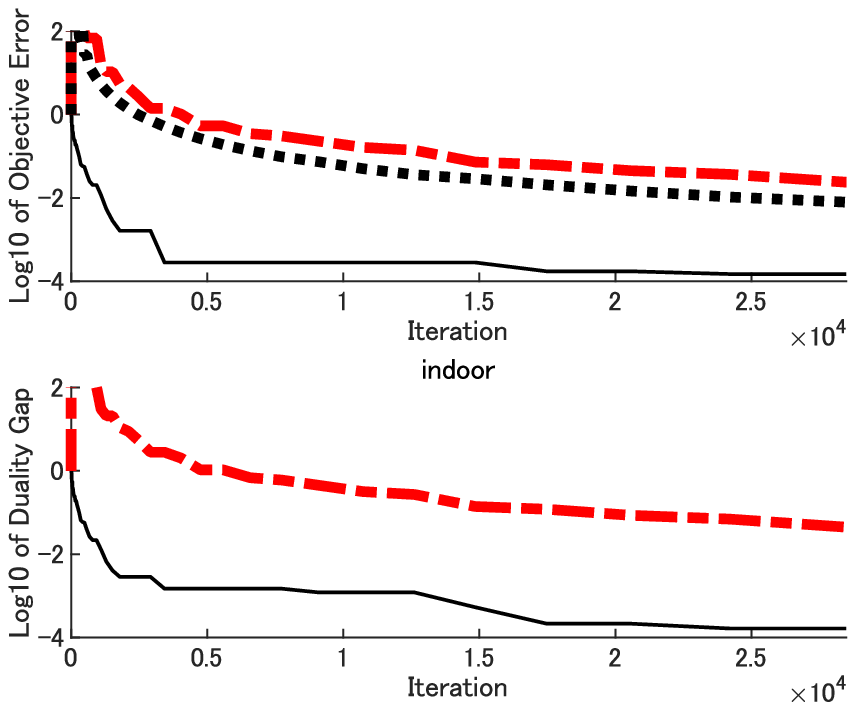}
    &
    \tsincgra{0.3}{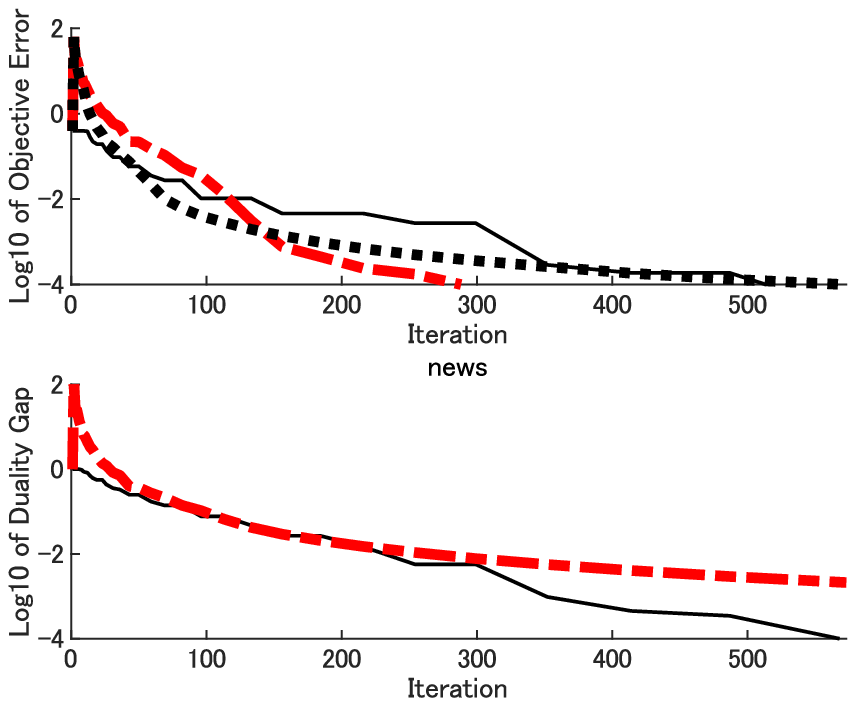}
    &
    \\
    \\
\end{tabular}
}
    \caption{Convergence behaviors for minimizing the empirical risk without the Moreau envelope. Regularization parameter is set to $\lambda=10^{0}/n$. 
      \label{fig:iter-vs-gap-fw}}
\end{center}
\end{figure*}
\begin{figure*}[t!]
  \begin{center}
    {\footnotesize
  \begin{tabular}{lll}
    (a) Caltech101 & (b) CUB200 & (c) Flower102
    \\
    \tsincgra{0.3}{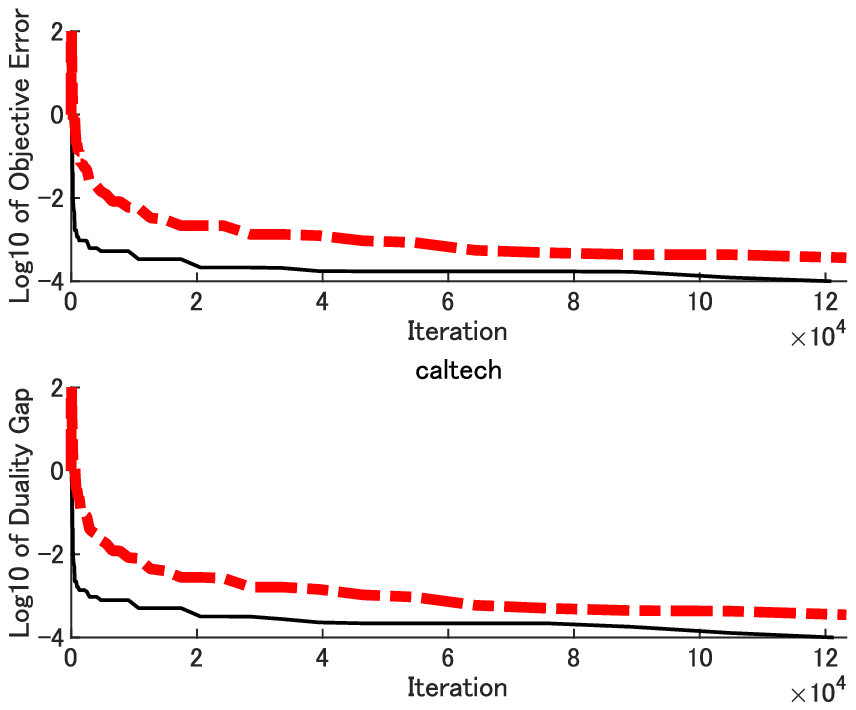}
    &
    \tsincgra{0.3}{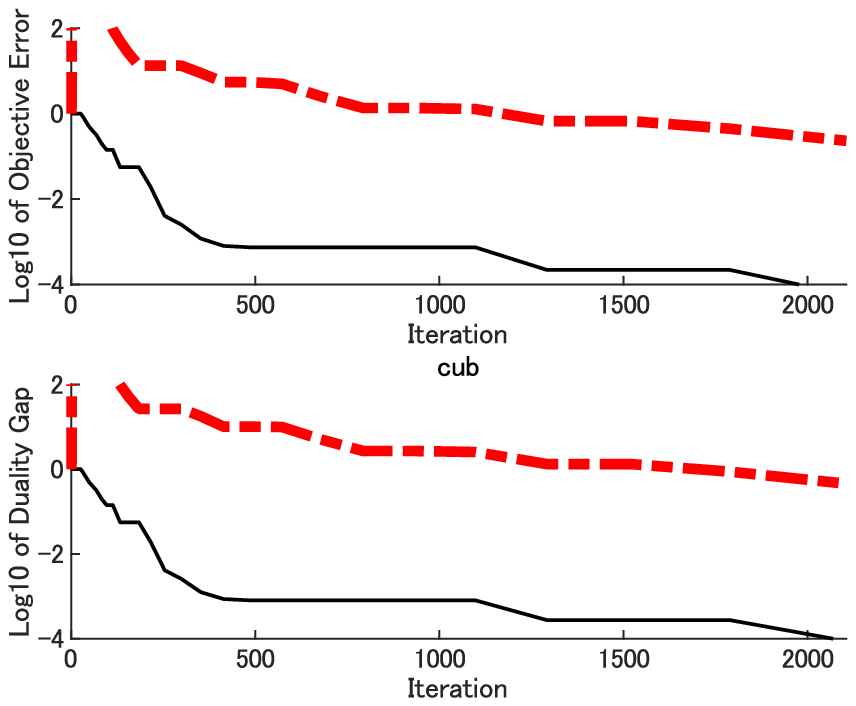}
    &
    \tsincgra{0.3}{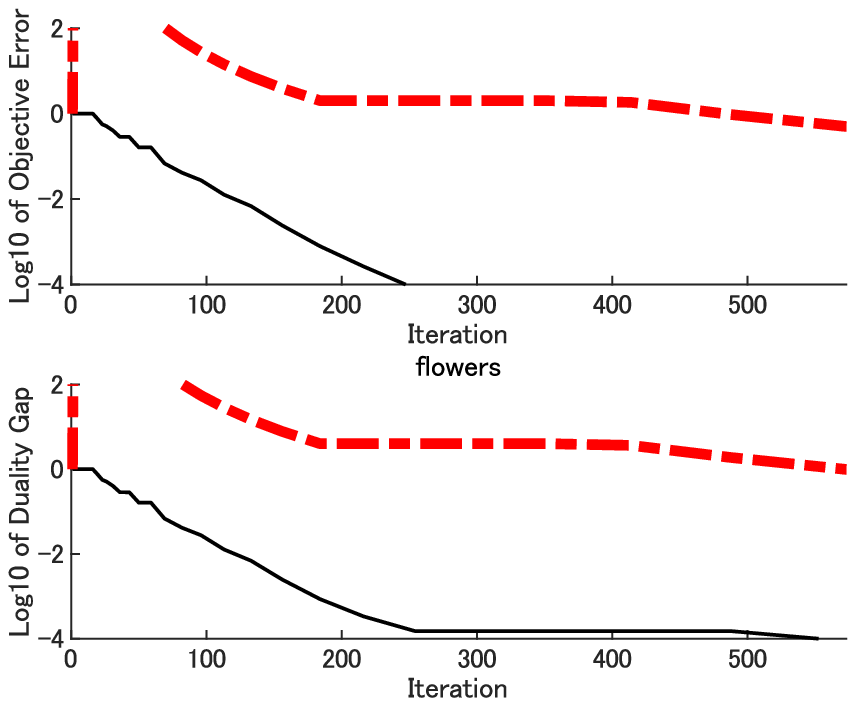}    
    \\
    \\
    (d) Indoor67 & (e) News20 & 
    \\
    \tsincgra{0.3}{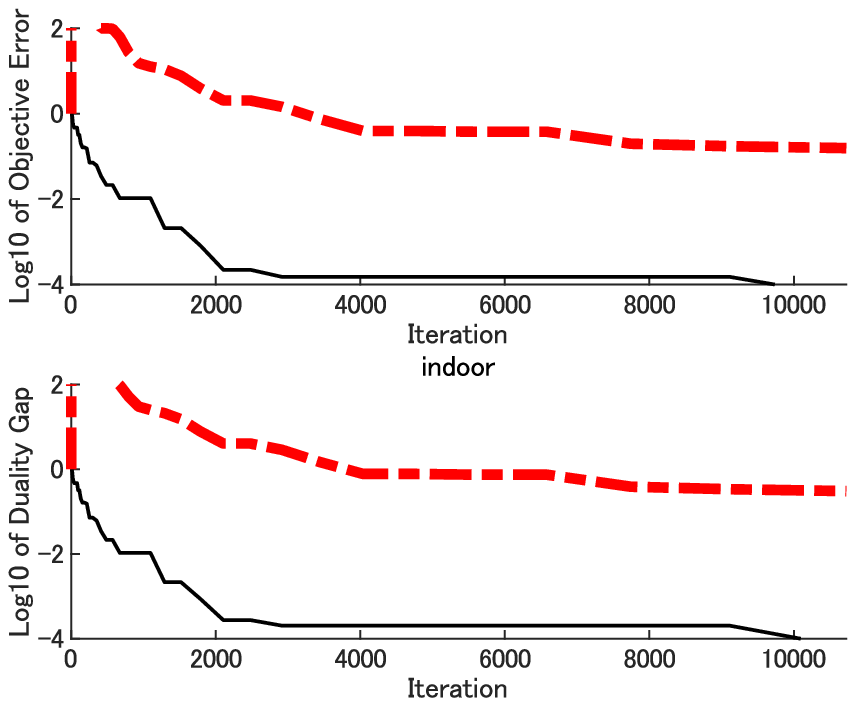}
    &
    \tsincgra{0.3}{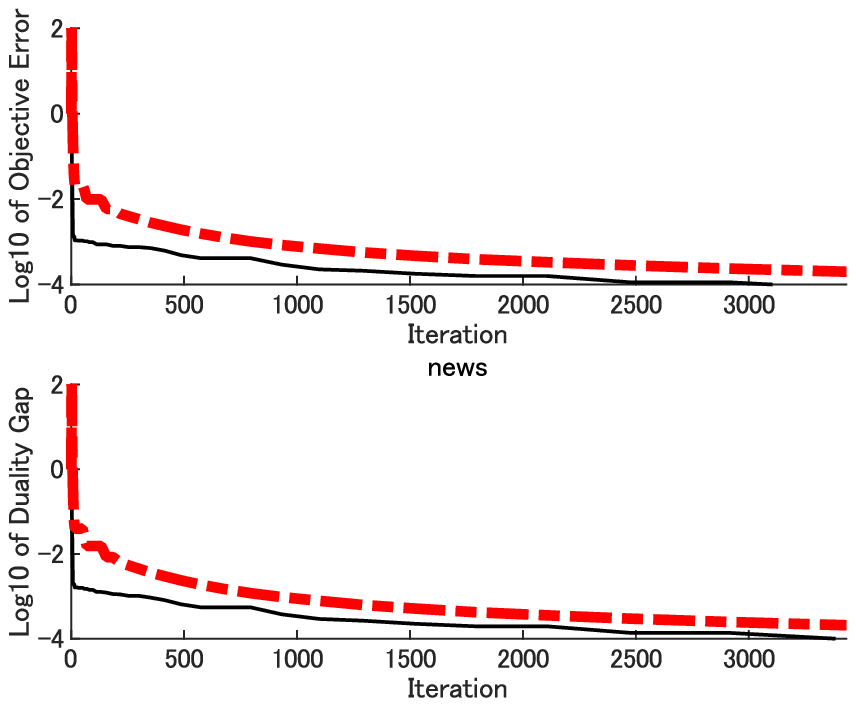}
    &
    \\
    \\
\end{tabular}
}
    \caption{Convergence behaviors for minimizing the empirical risk with the Moreau envelope. Regularization parameter is set to $\lambda=10^{0}/n$. 
      \label{fig:iter-vs-gap-smfw}}
\end{center}
\end{figure*}

\begin{table*}[t!]
  \centering
  \caption{Pattern recognition performance. \label{tab:pat}}
  \begin{tabular}{l}
    (a) Caltech101
    \\
\begin{tabular}{c|cccc}\hline
& Top-1 & Top-3 & Top-5 & Top-10 \\ \hline
PG & \underbar{0.469}(0.008) & 0.307(0.005)  & 0.257(0.005)  & 0.183(0.006) \\
StdFW   & \underbar{0.481}(0.022) & 0.317(0.032)  & 0.266(0.032)  & 0.192(0.035) \\
LSFW    & \textbf{\underbar{0.467}}(0.048)  & \textbf{\underbar{0.300}}(0.010)   & \textbf{\underbar{0.244}}(0.005)   & \textbf{\underbar{0.169}}(0.005)  \\
\hline \end{tabular}
\\ \\
(b) CUB200
\\
\begin{tabular}{c|cccc}\hline
& Top-1 & Top-3 & Top-5 & Top-10 \\ \hline
PG & 0.431(0.008) & 0.247(0.007)  & 0.184(0.004)  & 0.114(0.003) \\
StdFW   & 0.443(0.022) & 0.257(0.026)  & 0.196(0.026)  & 0.126(0.026) \\
LSFW    & \textbf{\underbar{0.415}}(0.006)  & \textbf{\underbar{0.231}}(0.007)   & \textbf{\underbar{0.166}}(0.006)   & \textbf{\underbar{0.098}}(0.004)  \\
\hline \end{tabular}
\\ \\
(c) Flower102
\\
\begin{tabular}{c|cccc}\hline
& Top-1 & Top-3 & Top-5 & Top-10 \\ \hline
PG & 0.219(0.009) & 0.107(0.010)  & 0.075(0.008)  & 0.041(0.004) \\
StdFW   & \underbar{0.220}(0.014) & \underbar{0.108}(0.013)  & \underbar{0.076}(0.011)  & \underbar{0.041}(0.009) \\
LSFW    & \textbf{\underbar{0.211}}(0.010)  & \textbf{\underbar{0.104}}(0.011)   & \textbf{\underbar{0.071}}(0.010)   & \textbf{\underbar{0.037}}(0.004)  \\
\hline \end{tabular}
\\ \\
(d) Indoor67
\\
\begin{tabular}{c|cccc}\hline
& Top-1 & Top-3 & Top-5 & Top-10 \\ \hline
PG & 0.319(0.005) & 0.139(0.003)  & 0.092(0.003)  & 0.051(0.002) \\
StdFW   & 0.323(0.016) & 0.141(0.009)  & 0.098(0.007)  & 0.058(0.008) \\
LSFW    & \textbf{\underbar{0.303}}(0.004)  & \textbf{\underbar{0.119}}(0.004)   & \textbf{\underbar{0.071}}(0.003)   & \textbf{\underbar{0.033}}(0.002)  \\
\hline \end{tabular}
\\ \\
(e) News20
\\
\begin{tabular}{c|cccc}\hline
& Top-1 & Top-3 & Top-5 & Top-10 \\ \hline
PG & 0.348(0.003) & 0.142(0.003)  & 0.083(0.003)  & 0.031(0.002) \\
StdFW   & 0.348(0.003) & 0.143(0.004)  & 0.084(0.005)  & 0.052(0.027) \\
LSFW    & \textbf{\underbar{0.339}}(0.004)  & \textbf{\underbar{0.136}}(0.002)   & \textbf{\underbar{0.080}}(0.005)   & \textbf{\underbar{0.028}}(0.002)  \\
\hline \end{tabular}
\\
  \end{tabular}
\end{table*}

\begin{table}[t!]
  \centering
  \caption{Computational times for one iteration. \label{tab:tmperiter}}
  \begin{tabular}{|c|ccc|}
    \hline
    & PG & StdFW & LSFW
    \\
    \hline
    Caltech101
    & 0.453 sec & 0.589 sec & 0.814 sec
    \\
    CUB200
    & 2.570 sec & 2.458 sec & 3.324 sec
    \\
    Flower102
    & 0.661 sec  & 0.649 sec  & 0.903 sec
    \\
    Indoor67
    & 3.040 sec  & 2.855 sec  & 3.919 sec
    \\
    News20 
    & 0.548 sec  & 0.436 sec  & 0.531 sec
    \\
    \hline
  \end{tabular}
\end{table}
\subsection{Extension to Moreau Envelope}
The Moreau envelope~\cite{Bertsekas99} is a trick often used for transforming a non-differentiable convex function into a smoothed function. For example, the Huber loss~\cite{Huber1964} and the smoothed hinge loss~\cite{Shalev-Shwartz2013jmlr}, widely used in machine learning, are, respectively, the Moreau envelopes of the absolute error and the hinge loss. Since it tends to take a shorter time to minimize a smooth objective function, the Moreau envelope is a useful technique to make machine learning efficient. The Moreau envelope of a convex loss function $\Phi(\cdot\,;\,y):\bR^{m}\to\bR$ is defined as
\begin{tsaligned}
  \Phi_{\text{m}}(\vs\,;\,y)
  :=
  \left(
  \Phi^{*}(\cdot\,;\,y)+\frac{\gamma_{\text{sm}}}{2}\lVert\cdot\rVert^{2}
  \right)^{*}(\vs\,;\,y)
\end{tsaligned}
where $\gamma_{\text{sm}}$ is a non-negative constant called the \emph{smoothing parameter}.
As long as $\Phi(\cdot\,;\,y)$ is a convex function,
its Moreau envelope $\Phi_{\text{m}}(\cdot\,;\,y)$ is ensured to be
\emph{$(1/\gamma_{\text{sm}})$-smooth}. To minimize the regularized empirical risk 
\begin{tsaligned}
  P_{\text{m}}(\vw)
  :=
  \frac{\lambda}{2}
  \lVert\vw\rVert^{2}
  +
  \frac{1}{n}
  \sum_{i=1}^{n}
  \Phi_{\text{m}}(\vPsi(\x_{i})^\top\vw\,;\,y_{i}), 
\end{tsaligned}
we now consider applying again the Frank-Wolfe method 
to maximization of the Fenchel dual 
\begin{tsaligned}
  D_{\text{m}}(\vA)
  :=
  -\frac{\lambda}{2}
  \lVert\vw(\vA)\rVert^{2}
  -
  \frac{1}{n}
  \sum_{i=1}^{n}
  \Phi_{\text{m}}^{*}(-\valph_{i}\,;\,y_{i}). 
\end{tsaligned}
Notice that the Moreau envelope of the loss function $\Phi(\cdot\,;\,y)$ is no longer mdos-type even if $\Phi(\cdot\,;\,y)$ is mdos-type, implying that the optimization algorithm presented in the previous section cannot be applied directly to the Moreau envelope. 
We obtained the following result: 
\begin{theorem-waku}\label{thm:fwsm}
  Consider applying the Frank-Wolfe algorithm to the problem
  for maximizing $D_{\text{m}}(\vA)$ with respect to $\vA$.
  Assume that the loss function $\Phi(\cdot\,;\,y)$ to be mdos-type. 
  Then, both the direction finding step and the line search step
  are expressed in a closed form even if $\gamma_{\text{sm}}>0$. 
\end{theorem-waku}
See Section~\ref{sec:proof-for-thm-fwsm} for the proof of Theorem~\ref{thm:fwsm}. 
The update rules of the direction finding step and the line search step
are described below.
Let
\begin{tsaligned}
  \tilde{\vs}_{i}^{(t-1)}:=\vPsi(\x_{i})^\top\vw(\vA^{(t-1)})-\gamma_{\text{sm}}\valph_{i}^{(t-1)}. 
\end{tsaligned}
The update rule of the direction finding step \eqref{eq:small-lp-sol-fwgeneral}
is replaced to 
\begin{tsaligned}\label{eq:small-lp-sol-fwsm}
  \vu_{i}^{(t-1)}\in-\partial\Phi
  \left(
  \tilde{\vs}_{i}^{(t-1)}
  \,;\,y_{i}
  \right)
\end{tsaligned}
Note that $\Phi(\cdot;y)$ in \eqref{eq:small-lp-sol-fwsm}
is the mdos loss function, not its Moreau envelope.
The expression of
$\hat{\gamma}^{(t-1)}$
for the line search step $\gamma^{(t-1)}=\max(0,\min(1,\hat{\gamma}^{(t-1)}))$
is replaced to 
\begin{tsaligned}\label{eq:gamhat-fwsm}
  &\hat{\gamma}^{(t-1)}
  :=
  \frac{%
    \sum_{i=1}^{n}%
    \left<%
    \Delta\valph_{i}^{(t-1)}, 
    \ve_{y_{i}}-\tilde{\vs}_{i}^{(t-1)}%
    \right>%
  }{%
    \lambda n
    \lVert\vw(\Delta\vA^{(t-1)})\rVert^{2}
    +
    \frac{\gamma_{\text{sm}}}{\lambda n}
    \lVert\Delta\vA^{(t-1)}\rVert^{2}
  }. 
\end{tsaligned}
The detailed derivations are described in
the proof of Theorem~\ref{thm:fwsm} in Appendix.
The procedure is summarized in Algorithm~\ref{algo:smfw}.
\begin{algorithm}[t!]
  \caption{
    Frank-Wolfe algorithm for minimizing a risk based on Moreau envelope. 
    \label{algo:smfw}. }
  \begin{algorithmic}[1]
    \STATE{$\vA^{(0)}\in\text{dom}(-D)$;}
    \FOR{$t:=1$ \textbf{to} $T$}
      \STATE\customlabel{line:smfw:vw-upd}{3}
      {$\vw^{(t-1)}:=\frac{1}{\lambda n}\sum_{i=1}^{n}\vPsi(\x_{i})\valph_{i}^{(t-1)}$;}
      \FOR{$i\in[n]$}
        \STATE\customlabel{line:smfw:vstil-upd}{5}
        {$\tilde{\vs}_{i}^{(t-1)}:=\vPsi(\x_{i})^\top\vw^{(t-1)}-\gamma_{\text{sm}}\valph_{i}^{(t-1)}$;}
        \STATE\customlabel{line:smfw:vu-upd}{6}
        {$\vu_{i}^{(t-1)}\in-\partial\Phi \left( \tilde{\vs}_{i}^{(t-1)}\,;\,y_{i}\right)$;}
        \STATE\customlabel{line:smfw:Del-alph-upd}{7}
        {$\Delta\valph^{(t-1)}_{i}:=\vu^{(t-1)}_{i}-\valph^{(t-1)}_{i}$;}
      \ENDFOR
      \STATE\customlabel{line:smfw:Del-vw-upd}{9}
      {$\Delta \vw^{(t-1)}:=\frac{1}{\lambda n}\sum_{i=1}^{n}\vPsi(\x_{i})\Delta\valph_{i}^{(t-1)}$;}
      \STATE\customlabel{line:smfw:hat-gam-upd}{10}
      {$\hat{\gamma}^{(t-1)}
      :=
      \frac{%
        \sum_{i=1}^{n}%
        \left<%
        \Delta\valph_{i}^{(t-1)}, 
        \ve_{y_{i}}-\tilde{\vs}_{i}^{(t-1)}%
        \right>%
      }{%
        \lambda n
        \lVert\Delta \vw^{(t-1)}\rVert^{2}
        +
        \frac{\gamma_{\text{sm}}}{\lambda n}
        \lVert\Delta\vA^{(t-1)}\rVert^{2}
      }$;}
      \STATE\customlabel{line:smfw:gam-upd}{11}
      {$\gamma^{(t-1)}:=\max(0,\min(1,\hat{\gamma}^{(t-1)}))$; }
      \STATE\customlabel{line:smfw:amat-upd}{12}
      {$\vA^{(t)} := \vA^{(t-1)}+\gamma^{(t-1)}\Delta\vA^{(t-1)}$;}
    \ENDFOR
  \end{algorithmic}
\end{algorithm}

In summary, it turns out that the two steps in each iteration of Frank-Wolfe method are expressed in a closed form not only for the mdos-type loss function but also for its Moreau envelope.

\subsection{Time Complexity}
We analyze the time complexity of the Frank Wolfe algorithm presented in Algorithm~\ref{algo:smfw}. The middle column in Table~\ref{tab:compcost} shows the time complexity consumed in each line.
Line~\ref{line:smfw:vstil-upd} and Line~\ref{line:smfw:Del-alph-upd}, respectively, require $O(md)$ and $O(m)$ for each $i\in[n]$, which take $O(md)\times n$ and $O(m)\times n$ to compute $n$ vectors of $\tilde{\vs}_{i}^{(t-1)}$ and $\Delta\valph^{(t-1)}_{i}$. Line~\ref{line:smfw:vu-upd} contains computation of a subgradient of the loss function. The time complexity for this line depends on the definition of the loss function. The max hinge loss requires $O(m)$ computation for Line~\ref{line:smfw:vu-upd}. The top-$k$ hinge loss, the Usunier loss, and their weighted generalizations consume $O(m\log m)$ computation for the line. 
\begin{table}
  \caption{Time complexity of each line in Algorithm~\ref{algo:smfw}. \label{tab:compcost}}
\begin{center}
  \begin{tabular}{lll}
    \hline
    Line Number &
    General case &
    A special case
    \\
    \hline
    \hline
    Line~\ref{line:smfw:vw-upd}
    &
    $O(mnd)$
    &
    $O(nd)$
    \\
    Line~\ref{line:smfw:vstil-upd}
    &
    $O(md)\times n$
    &
    $O(d)\times n$
    \\
    Line~\ref{line:smfw:vu-upd}
    &
    Depends on $\Phi$
    &
    Depends on $\Phi$
    \\
    Line~\ref{line:smfw:Del-alph-upd}
    &
    $O(m)\times n$
    &
    $O(m)\times n$
    \\
    Line~\ref{line:smfw:Del-vw-upd}
    &
    $O(mnd)$
    &
    $O(nd)$
    \\
    Line~\ref{line:smfw:hat-gam-upd}
    &
    $O(mn+d)$
    &
    $O(mn+d)$
    \\
    Line~\ref{line:smfw:amat-upd}
    &
    $O(mn)$
    &
    $O(mn)$
    \\
    \hline
  \end{tabular}
\end{center}
\end{table}

In the case that the feature extractor is
$\vpsi(\x,j)=\ve_{j}\otimes\x$ with $\cX:=\bR^{d_{0}}$, 
the time complexity is improved compared to the general case.
In this case, $d=m d_{0}$. The update rule of the primal variable in Line~\ref{line:smfw:vw-upd} in Algorithm~\ref{algo:smfw} is rewritten as: 
\begin{tsaligned}
  \vw^{(t-1)}
  =
  \frac{1}{\lambda n}\sum_{i=1}^{n}\valph_{i}^{(t-1)}\otimes\x_{i}. 
\end{tsaligned}
Now we discuss how to update $\tilde{\vs}_{i}^{(t-1)}$ in Line~\ref{line:smfw:vstil-upd}. To compute the $j$th entry in the $m$-dimensional vector $\vPsi(\x_{i})^\top\vw^{(t-1)}$, we extract a sub-vector $(\ve_{j}^\top\otimes\vI_{d_{0}})\vw$ and take the inner-product between it and the input vector $\x_{i}$. Both extraction of $m$ sub-vectors and computation of $m$ inner-products takes $O(m)$ computational cost.
Those discussion is summarized in the third column of Table~\ref{tab:compcost}. 

\section{Empirical Evaluation}
In this section, we demonstrate the power of the proposed Frank-Wolfe algorithm in terms of the convergence speed and the pattern recognition performance.

\subsection{Convergence Speed}
To illustrate how rapidly the proposed optimization algorithm for empirical risk minimization are converged, we used five image datasets Caltech101 Silhouettes, CUB200, Flower102 and Indoor67 containing $n=6,339$, $6,033$, $2,040$ and $15,607$ images, respectively, and a text dataset News20 containing $15,935$ texts. The images or texts in each dataset are classified into $m=102$, $200$, $102$, $67$, $20$ categories, respectively. 
Deep neural structure named VGG16 was used to extract an input vector $\x\in\bR^{d}$ from each image for CUB200, Flower102, and Indoor67. 
We extract $d=4,096$ features from the deep networks for the three image datasets, respectively.  
Features for Caltech101 were the vectorization of the pixel intensities of $16\times 16$ gray scaled images.  
Singular value decomposition was performed to reduce 15,935 word counts in News20 to $d=1,024$ features.  
For the loss function, \eqref{eq:wu-def} was chosen with $\rho_{j}:=\max(0, 6-j)/15$. The regularization parameter is set to $\lambda=1/n$. The proposed optimization algorithm was compared with two methods: \texttt{PG} and \texttt{StdFW}. The method \texttt{PG} is the projected gradient algorithm~\cite{Shalev-Shwartz11a-pegasos}. In each iteration, \texttt{PG} updates the primal variable to the descent direction and projects it onto a ball to suppress the norm of the gradient vector. The sublinear convergence is ensured for Lipschitz continuous loss functions. The method \texttt{StdFW} is an alternative to the proposed Frank-Wolfe algorithm. In every iteration of the proposed algorithm, the direction finding step is followed by the line search step that finds the optimal ratio, say $\gamma_{t}$, for mixing the previous point with the new point computed at the direction finding step. Theoretically, the sublinear convergence is guaranteed even if the ratio is pre-scheduled with $\gamma_{t}=2/(t+1)$. The method \texttt{StdFW} denotes the Frank-Wolfe using the pre-scheduled $\gamma_{t}$, while the proposed Frank-Wolfe is referred to as the line search Frank-Wolfe abbreviated with \texttt{LSFW}. 

Figure~\ref{fig:iter-vs-gap-fw} have five panels, each of which is for one of the five datasets. Each panel contains two sub-panels. The upper and lower sub-panels, respectively, show the objective errors and the duality gaps against the number of iterations, where the objective error and the duality gap at $t$-th iteration are defined as $P(\vw^{(t)})-P(\vw_{\star})$ and $P(\vw^{(t)})-D(\vA^{(t)})$, respectively, where $\vw^{(t)}$ and $\vA^{(t)}$ are the values of the primal and the dual variables at $t$-th iteration. For \texttt{StdFW} and \texttt{LSFW}, the value of the primal variable is recovered by $\vw^{(t)}:=\vw(\vA^{(t)})$ for $t\in\bN$. Since it is impossible to know the exact value of the optimal solution $\vw_{\star}$, the value of $\vw_{\star}$ is approximated by $\vw_{\star}\approx \vw(\vA^{(t')})$ in this experiments, where $\vA^{(t')}$ is obtained by iterating the Frank-Wolfe until reaching $P(\vw^{(t)})-D(\vA^{(t)})<10^{-5}$. From Figure~\ref{fig:iter-vs-gap-fw}, it can be observed that \texttt{LSFW} converges much faster than \texttt{PG}. 
For CUB200, Flower102 and Indoor67, the convergence of LSFW was much faster than those of the other two methods, whereas the convergence speed among three methods were similar for Caltech101 and News20. 
A property of the three datasets, CUB200, Flower102 and Indoor67, differs from that of the two datasets, Caltech101 and News20. The property is the number of dimensions of feature vectors. Feature vectors in CUB200, Flower102 and Indoor67 have a higher dimension than these in Caltech101 and News20. High-dimensional features tend to make the dual objective function more strongly concave. The authors conjecture that the difference in the number of dimensions yields the differences of the convergence behaviors. 

We then applied the Moreau envelope to the loss function with $\gamma_{\text{sm}}=0.01$. The convergence behaviors were changed as shown in Figure ~\ref{fig:iter-vs-gap-smfw}
. It can be shown that the negative dual function is strongly convex with the coefficient $\gamma_{\text{sm}}/n$ whatever training data are given. Indeed, the Moreau envelope make the convergences of LSFW for Caltech101 and News20 faster, although the convergences for the other four datasets were not accelerated. An explanation of this phenomenon may be that the negative dual objectives for CUB200, Flower102, and Indoor67 are already strongly convex even without the Moreau envelope. 

Table~\ref{tab:tmperiter} shows the computational times for each iteration of the three optimization algorithms. The running times for PG and StdFW are similar. Compared to StdFW, LSFW has to take more computation to perform line search. The computational times of LSFW do not exceed 1.5 times of the times of StdFW for one iteration. By combining the experimental results for the running time for one iteration with the objective errors against the number of iterations, it can be concluded that LSFW can achieve accurate solutions with much smaller computational times. 
\subsection{Pattern Recognition Performance}
We examined the pattern recognition performance on the five datasets used for the convergence experiments. For each dataset, 50\% of data were randomly picked. Each of three optimization algorithms was applied to the picked data to train a multi-category classifier. The rest of the data were used for testing the generalization performance of pattern recognition. Each optimization algorithm was implemented for 1,000 iterations. Cross-validation was performed to determine the value of the regularization constant $\lambda$. Top-$1$, top-$3$, top-$5$ and top-$10$ error ratios were used for assessing the generalization performance. Lower value indicates better performance. The above procedures were preformed 20 times, and the averages and the standard deviations of the performance measures across the 20 trials are reported in Table~\ref{tab:pat}. The bold-faced figures indicate the best performance. The underlined figures have no significant difference from the best performance, where the significance is based on the one-sample t-test. For all datasets, LSFW achieved the smallest top-$k$ error. Most of the error ratios of LSFW were significantly smaller than those of StdFW and PG. That might be because LSFW successfully produces sufficiently accurate solutions for training within 1,000 iterations whereas the other two could not. 
\section{Conclusions}
In this paper, we presented a new Frank-Wolfe algorithm that can be applied to the mdos-type learning machines. The mdos type is a class introduced newly in this study to analyze, in a unified fashion, a wide variety of loss functions originated from the max-hinge loss. The sublinear convergence of the Frank-Wolfe algorithm is ensured if both the direction finding step and the line search step are exactly implemented. We discovered that, if the Frank-Wolfe is applied to the Fenchel dual of the regularized empirical risk function, closed form solutions exist both for the two steps. Since low-dimensional feature vectors often slow down the minimization algorithms including the Frank-Wolfe, the loss function is often replaced to its Moreau envelope. However, the replaced loss function is no longer mdos-type, meaning that the proposed Frank-Wolfe cannot be applied directly. Nevertheless, we found a technique to reuse the proposed Frank-Wolfe again for the Moreau envelope of the loss function. We carried out experiments to empirically show that our algorithm converges faster and achieves a better pattern recognition performance compared to the existing methods. 






\bibliographystyle{plain}


\newpage
\appendix
\onecolumn
\section{Proof for Theorem~\ref{thm:five-losses-mdos-typ}}
\label{sec:proof-for-thm-mdos}
Proposition~2 and Proposition~5 in \cite{lapin-nips2015}, respectively,
show that 
the two loss functions $\Phi_{\text{utk}}(\cdot;y)$ and $\Phi_{\text{uu}}(\cdot;y)$ are mdos-type. 
The proof for $\Phi_{\text{sh}}(\cdot;y)$ is straightforward
because the expression of $\Phi_{\text{sh}}(\cdot;y)$ is similar
to those of $\Phi_{\text{uu}}(\cdot;y)$ and $\Phi_{\text{utk}}(\cdot;y)$
with $k=1$.
Kato \& Hirohashi has already shown in (15) of \cite{KatHir-acml19a}
that $\Phi_{\text{wtk}}(\cdot;y)$ is mdos-type. 
In what follows, we shall show that $\Phi_{\text{wu}}(\cdot;y)$
is mdos-type, for which it suffices to prove the following equation: 
\begin{tsaligned}\label{eq:smlfied-wu-is-maxdot}
  \sum_{\ell=1}^{|\cK|}
  \max\left\{
  0,
  \rho^{\prime}_{\ell} \left<\1,\x_{\pi(1:k_{\ell})}\right>
  \right\}
  =
  \max_{\vbeta\in\cB_{\text{wu}}}
  \left< \vbeta, \x \right>. 
\end{tsaligned}
The left hand side can be rearranged as: 
\begin{tsaligned}
&
\text{LHS of \eqref{eq:smlfied-wu-is-maxdot}}
=
\min\left\{
\sum_{j=1}^{m}h_{\pi(j)}\rho_{j}
\,\middle|\,
\vh \ge \0_{m},\, \vh \ge \x
\right\}
\\
&=
\min\left\{
\sum_{\ell=1}^{L}
\left<\1,\vh_{\vpi(1:k_{\ell})}\right>
\rho'_{\ell}
\,\middle|\,
\vh \ge \0_{m},\, \vh \ge \x
\right\}
\\
&=
\min\Bigg\{
\sum_{\ell=1}^{L}
\left(
k_{\ell}t_{\ell}
+
\left<\1,
\max(\0,\vh-t_{\ell}\1)
\right>
\right)
\rho'_{\ell}
\,\Bigg|
\vh \ge \0_{m},\, \vh \ge \x, \,
\vt \ge \0_{L}
\Bigg\}
\\
&=
\min\Bigg\{
\sum_{\ell=1}^{L}
\left(
k_{\ell}t_{\ell}
+
\left<\1,
\vq_{\ell}
\right>
\right)
\rho'_{\ell}
\,\Bigg|\,
\vh \ge \0_{m},\, \vh \ge \x,
\vt \ge \0_{L},\,
\ell\in [L], \,
\vq_{\ell}\ge 0_{m}, \,
\vq_{\ell}\ge \vh-t_{\ell}
\Bigg\}. 
\end{tsaligned}
To find an analytical solution to the above minimization problem,
a non-negative Lagrangian multiplier vector is
introduced for each constraint: 
\begin{center}
  \begin{tabular}{lll}
    &
    $\veta\in\bR_{+}^{m}$
    &
    $\text{for the constraint }\vh \ge \0$,
    \\
    &
    $\vbeta\in\bR_{+}^{m}$
    &
    $\text{for the constraint }\vh \ge \x$,
    \\
    $\ell\in[L]$, 
    &
    $\vmu_{\ell}\in\bR_{+}^{m}$
    &
    $\text{for the constraint }\vq_{\ell} \ge \0$
    \\
    $\ell\in[L]$, 
    &
    $\vlam_{\ell}\in\bR_{+}^{m}$
    &
    $\text{for the constraint }\vq_{\ell} \ge \vh-t_{\ell}\1$, and
    \\
    &
    $\vtau\in\bR_{+}^{L}$
    &
    $\text{for the constraint }\vt\ge\0$. 
  \end{tabular}
\end{center}
Let 
\begin{tsaligned}
  &\vQ := \left[\vq_{1},\dots,\vq_{L}\right],
  \quad
  \vM := \left[\vmu_{1},\dots,\vmu_{L}\right],
  \\
  &
  \quad
  \text{and}
  \quad
  \vLam := \left[\vlam_{1},\dots,\vlam_{L}\right]. 
\end{tsaligned}
The Lagrangian function is expressed as
\begin{tsaligned}
  &
  \cL_{\text{wu}}(\vh,\vt,\vQ,\veta,\vbeta,\vM,\vLam,\vtau)
  \\
  &:=
  \left<\vk\odot\vt+\vQ^\top\1,\vrho'\right>
  -
  \left<\veta,\vh\right>
  +
  \left<\vbeta,\x-\vh\right>
  \\
  &\qquad
  -
  \left<\vM,\vQ\right>
  +
  \left<\vLam,\vh\1^\top-\1\vt^\top-\vQ\right>
  -
  \left<\vtau,\vt\right>
  \\
  &=
  \left<\vt,\vrho'\odot\vk - \vLam^\top\1 - \vtau\right>
  +
  \left<\vQ, \1\vrho^{\prime\top}-\vLam-\vM\right>
  \\
  &\qquad
  +
  \left<\vh,\vLam\1-\vbeta-\veta\right>
  +
  \left<\x,\vbeta\right>. 
\end{tsaligned}
KKT conditions lead to
\begin{tsaligned}
  &\vrho'\odot\vk - \vLam^\top\1 - \vtau = \0,
  \quad
  \1\vrho^{\prime\top}-\vLam-\vM = \vO,
  \\
  &
  \quad
  \text{and}
  \quad
  \vLam\1-\vbeta-\veta = \0.  
\end{tsaligned}
Eliminating $\vtau$, $\vM$ and $\veta$,
the above conditions can be rewritten as
\begin{tsaligned}
  \vrho'\odot\vk \ge \vLam^\top\1, 
  \quad
  \1\vrho^{\prime\top} \ge \vLam, 
  \quad\text{and}
  \quad
  \vLam\1\ge \vbeta,  
\end{tsaligned}
Hence, we conclude that
\begin{tsaligned}
  &\text{LHS of \eqref{eq:smlfied-wu-is-maxdot}}
  =
\min_{\vh,\vt,\vQ}\max_{\veta,\vbeta,\vM,\vLam,\vtau}
\cL_{\text{wu}}(\vh,\vt,\vQ,\veta,\vbeta,\vM,\vLam,\vtau)
\\
&=
\max
\Big\{
\left<\x,\vbeta\right>
\,\Big|\,
  \vrho'\odot\vk \ge \vLam^\top\1, \,
  \1\vrho^{\prime\top} \ge \vLam, \,
  \vLam\1\ge \vbeta  
  \Big\}
  \\
  &
  =
  \text{RHS of \eqref{eq:smlfied-wu-is-maxdot}}. 
\end{tsaligned}
\qed
\section{Proof for Theorem~\ref{thm:fw}}
\label{sec:proof-for-thm-fw}
We first show that the direction finding step
can be expressed in a closed form as in \eqref{eq:small-lp-sol-fwgeneral},
followed by showing that the solution to the
line search step is
$\gamma^{(t-1)}=\max(0,\min(1,\hat{\gamma}^{(t-1)}))$. 

\subsection{Proof for Direction Finding Step}
We use Lemma~1 of \cite{KatHir-acml19a} to
rearrange the dual objective function as: 
\begin{tsaligned}\label{eq:D-mdos}
  D(\vA)
  :=
  -
  \frac{\lambda}{2}\lVert\vw(\vA)\rVert^{2}
  +
  \frac{1}{n}\sum_{i=1}^{n}\left<\ve_{y_{i}},\valph_{i}\right>. 
\end{tsaligned}
The derivative with respect to $\valph_{i}$ is obtained as
\begin{tsaligned}\label{eq:D-deriv-wrt-alphi}
  \frac{\partial D(\vA)}{\partial \valph_{i}}
  =
  \frac{1}{n}
  \left(
  \ve_{y_{i}} - \vPsi(\x_{i})^\top\vw(\vA^{(t-1)})
  \right). 
\end{tsaligned}
The effective domain of $-D$ is the product space of
the feasible region of each column: 
\begin{tsaligned}
  \text{dom}(-D)
  =
  -\prod_{i=1}^{n}\text{dom}\Phi^{*}(\cdot\,;\,y_{i}), 
\end{tsaligned}
This fact allows us to decompose the $m\cdot n$-variable
linear programming problem to $n$ smaller problems: 
\begin{tsaligned}\label{eq:small-lp-fw}
  \vu_{i}^{(t-1)}
  \in
  \argmin_{\vu_{i}\in-\text{dom}\Phi^{*}(\cdot\,;\,y_{i})}
  \left<
  \vW(\vA^{(t-1)})^\top\x_{i} - \ve_{y_{i}},
  \vu_{i}
  \right>. 
\end{tsaligned}
Applying Lemma~3 of \cite{KatHir-acml19a},
an optimal solution to each of
the $n$ linear programming problems
can be expressed as \eqref{eq:small-lp-sol-fwgeneral}.
\qed

\subsection{Proof for Line Search Step}
Let $\vw^{(t-1)}:=\vw(\vA)$. We shall use
\begin{tsaligned}
  \lVert
  \vw(\gamma \Delta\vA)
  \rVert^{2}
  =
  \lVert
  \vw(\Delta\vA)
  \rVert^{2}
  \gamma^{2}
\end{tsaligned}
and
\begin{tsaligned}
  \left<
  \vw^{(t - 1)},\vw(\Delta\vA)
  \right>
  =
  \frac{1}{\lambda n}
  \left<
  \vw^{(t - 1)},
  \sum_{j=1}^{n}\vPsi(\x_{j})\Delta\valph_{j}
  \right>
  =
  \frac{1}{\lambda n}
  \sum_{j=1}^{n}
  \left<
  \vPsi(\x_{j})^\top\vw^{(t - 1)},
  \Delta\valph_{j}
  \right>
\end{tsaligned}
to obtain
\begin{tsaligned}
  & \frac{1}{\lambda}
  \left(
  D(\vA+\gamma\Delta\vA)
  -
  D(\vA)
  \right)
  \\
  &=
  -
  \frac{1}{2}
  \lVert
  \vw^{(t-1)}
  +
  \vw(\gamma \Delta\vA)
  \rVert^{2}
  +
  \frac{1}{2}  
  \lVert
  \vw^{(t-1)}
  \rVert^{2}
  \\
  &\quad
  +
  \frac{1}{\lambda n}
  \sum_{j=1}^{n}
  \left<\ve_{j},\valph_{j}+\gamma \Delta\valph_{j}\right>
  -
  \frac{1}{\lambda n}
  \sum_{j=1}^{n}
  \left<\ve_{j},\valph_{j}\right>
  \\
  &=
  -
  \frac{1}{2}
  \lVert
  \vw(\gamma \Delta\vA)
  \rVert^{2}
  -
  \left<\vw^{(t-1)}, \vw(\Delta\vA)\right>
  \gamma
  +
  \frac{1}{\lambda n}
  \sum_{j=1}^{n}
  \left<\ve_{j}, \Delta\valph_{j}\right>\gamma
  \\
  &=
  -
  \frac{1}{2}
  \lVert
  \vw(\Delta\vA)
  \rVert^{2}
  \gamma^{2}
  +
  \frac{1}{\lambda n}
  \sum_{j=1}^{n}
  \left<\ve_{j}-\vPsi(\x_{j})^\top\vw^{(t - 1)},
  \Delta\valph_{j}\right>\gamma. 
\end{tsaligned}
This concludes that the optimal solution
for line search is
$\gamma^{(t-1)}=\max(0,\min(1,\hat{\gamma}^{(t-1)}))$. 

\subsection{Derivation of \eqref{eq:D-deriv-wrt-alphi}}
The derivative of the second term in \eqref{eq:D-mdos}
with respect to $\valph_{i}$ is
\begin{tsaligned}\label{eq:D-mdos-1st-term-deriv}
  \frac{\partial}{\partial \valph_{i}}
  \sum_{j=1}^{n}\left<\ve_{y_{j}},\valph_{j}\right>
  =
  \ve_{y_{i}}. 
\end{tsaligned}
We now calculate the derivative of the first term 
in \eqref{eq:D-mdos}. Let 
\begin{tsaligned}
  \bar{\vw} := \frac{1}{\lambda n}\sum_{j\ne i}\vPsi(\x_{j})\valph_{j}
\end{tsaligned}
to obtain
\begin{tsaligned}
   \lVert\vw(\vA)\rVert^{2}
  =
  \lVert\bar{\vw} + \frac{1}{\lambda n}\vPsi(\x_{i})\valph_{i}\rVert^{2}
  =
  \lVert\bar{\vw}\rVert^{2}
  +
  \frac{2}{\lambda n}\left<\vPsi(\x_{i})^\top\bar{\vw},\valph_{i}\right>
  +
  \frac{1}{(\lambda n)^{2}}\lVert\vPsi(\x_{i})\valph_{i}\rVert^{2}. 
\end{tsaligned}
The derivative is
\begin{multline}\label{eq:D-mdos-2nd-term-deriv}
  \frac{\partial}{\partial \valph_{i}}\lVert\vw(\vA)\rVert^{2}
  =
  \frac{2}{\lambda n}\vPsi(\x_{i})^\top\bar{\vw}
  +
  \frac{2}{(\lambda n)^{2}}\vPsi(\x_{i})^\top\vPsi(\x_{i})\valph_{i}
  \\
  =
  \frac{2}{\lambda n}\vPsi(\x_{i})^\top
  \left(
  \bar{\vw}
  +
  \frac{1}{\lambda n}\vPsi(\x_{i})\valph_{i}
  \right)
  =
  \frac{2}{\lambda n}\vPsi(\x_{i})^\top
  \vw(\vA). 
\end{multline}
Combining
\eqref{eq:D-mdos-1st-term-deriv}
with
\eqref{eq:D-mdos-2nd-term-deriv},
we obtain \eqref{eq:D-deriv-wrt-alphi}. 
%

%
\section{Proof for Theorem~\ref{thm:fwsm}}
\label{sec:proof-for-thm-fwsm}
%
Let us consider an $(m+1)d\times m$ feature matrix for each
training example $(\x_{i},y_{i})$, defined as
\begin{tsaligned}
  \tilde{\vPsi}_{i}
  :=
  \begin{bmatrix}
    \vPsi(\x_{i})
    \\
    \sqrt{\lambda n\gamma_{\text{sm}}}\ve_{i}\otimes\vI_{m}
  \end{bmatrix} 
\end{tsaligned}
where $\ve_{i}$ is here an $n$-dimensional unit vector with
$i$-th entry one. 
From the $n$ feature matrices, we pose the 
regularized empirical risk~$\tilde{P}:\bR^{(m+1)d}\to\bR$ defined as
\begin{tsaligned}
  \tilde{P}(\tilde{\vw})
  :=
  \frac{\lambda}{2}\lVert\tilde{\vw}\rVert^{2}
  +
  \frac{1}{n}
  \sum_{i=1}^{n}
  \Phi( \tilde{\vPsi}_{i}^\top\vw\,;\,y_{i} )
\end{tsaligned}
where $\Phi(\cdot\,;\,y)$ is the mdos-type loss function.
To show Theorem~\ref{thm:fwsm}, we shall use the following lemma:
\begin{lemma-waku}\label{lem:Dm-is-dual-to-Ptil}
  Function $D_{\text{m}}$ is an Fenchel dual to $\tilde{P}$.
\end{lemma-waku}
The proof for Lemma~\ref{lem:Dm-is-dual-to-Ptil}
is given in Subsection~\ref{ss:proof-lem:Dm-is-dual-to-Ptil}. 
The proof for Theorem~\ref{thm:fwsm} is completed by deriving
\eqref{eq:small-lp-sol-fwsm}
and
\eqref{eq:gamhat-fwsm}.
Each derivation is given in 
Subsection~\ref{ss:deriv-eq:small-lp-sol-fwsm}
and
Subsection~\ref{ss:deriv-eq:gamhat-fwsm}, 
respectively. 

\subsection{Derivation of \eqref{eq:small-lp-sol-fwsm}}
\label{ss:deriv-eq:small-lp-sol-fwsm}
Define $\tilde{\vw}:\bR^{m\times n}\to\bR^{(m+1)d}$ as
\begin{tsaligned}
  \tilde{\vw}(\vA)
  :=
  \frac{1}{n}
  \sum_{i=1}^{n}
  \tilde{\vPsi}_{i}\valph_{i}. 
\end{tsaligned}
Observe that 
\begin{tsaligned}
\tilde{\vw}(\vA)
  =
  \begin{bmatrix}
    \frac{1}{\lambda n}
    \sum_{i=1}^{n}\vPsi(\x_{i})\valph_{i}
    \\
    \frac{\sqrt{\lambda n\gamma_{\text{sm}}}}{\lambda n}
    \sum_{i=1}^{n}(\ve_{i}\otimes\vI_{m})\valph_{i}
  \end{bmatrix}
  =
  \begin{bmatrix}
    \vw(\vA)
    \\
    \sqrt{\frac{\gamma_{\text{sm}}}{\lambda n}}\text{vec}(\vA)
  \end{bmatrix}
\end{tsaligned}
where $\text{vec}(\vA)$ is the vectorization of an $m\times n$ matrix $\vA$.
From Lemma~\ref{lem:Dm-is-dual-to-Ptil}, the direction finding step
in the Frank-Wolfe algorithm for maximizing $D_{\text{m}}$
can be written as
\begin{tsaligned}\label{eq:01-deriv-small-lp-sol-fwsm}
  \vu_{i}^{(t-1)}\in-\partial\Phi
  \left(
  \tilde{\vPsi}_{i}^\top\tilde{\vw}(\vA^{(t-1)})
  \,;\,y_{i}
  \right). 
\end{tsaligned}
The prediction score can be rearranged as
\begin{multline}\label{eq:02-deriv-small-lp-sol-fwsm}
  \tilde{\vPsi}_{i}^\top\tilde{\vw}(\vA^{(t-1)})
  =
  \vPsi(\x_{i})^\top\vw(\vA^{(t-1)})
  +
  \sqrt{\lambda n\gamma_{\text{sm}}}
  \sqrt{\frac{\gamma_{\text{sm}}}{\lambda n}}
  \left(\ve_{i}^\top\otimes\vI_{m}\right)
  \text{vec}(\vA^{(t-1)})
  \\
  =
  \vPsi(\x_{i})^\top\vw(\vA^{(t-1)})
  +
  \gamma_{\text{sm}} \valph_{i}^{(t-1)}
  =
  \tilde{\vs}_{i}^{(t-1)}. 
\end{multline}
Combining \eqref{eq:01-deriv-small-lp-sol-fwsm}
with \eqref{eq:02-deriv-small-lp-sol-fwsm},
the direction finding step can be obtained as 
\eqref{eq:small-lp-sol-fwsm}. 

\subsection{Derivation of \eqref{eq:gamhat-fwsm}}
\label{ss:deriv-eq:gamhat-fwsm}
From discussion in Section~\ref{s:fw}, the line search step
computes $\gamma^{(t-1)}$ by clipping
$\hat{\gamma}^{(t-1)}$ outside the interval of $[0,1]$
where
\begin{tsaligned}\label{eq:01-deriv-gamhat-fmsm}
  \hat{\gamma}^{(t-1)}
  =
  \frac{%
    \sum_{i=1}^{n}%
    \left<%
    \Delta\valph_{i}^{(t-1)}, 
    \ve_{y_{i}}-\tilde{\vPsi}_{i}^\top\tilde{\vw}(\vA^{(t-1)})%
    \right>%
  }{%
    \lambda n\lVert\tilde{\vw}(\Delta\vA^{(t-1)})\rVert^{2}
  }. 
\end{tsaligned}
The square of the norm of
$\tilde{\vw}(\Delta\vA^{(t-1)})$ is given by
\begin{tsaligned}\label{eq:02-deriv-gamhat-fmsm}
  \lVert\tilde{\vw}(\Delta\vA^{(t-1)})\rVert^{2}
  =
  \lVert\vw(\Delta\vA^{(t-1)})\rVert^{2}
  +
  \frac{\gamma_{\text{sm}}}{\lambda n}
  \lVert\Delta\vA^{(t-1)}\rVert^{2}_{\text{F}}. 
\end{tsaligned}
Equation \eqref{eq:gamhat-fwsm} is established
by substituting
\eqref{eq:02-deriv-small-lp-sol-fwsm}
and
\eqref{eq:02-deriv-gamhat-fmsm}
into
\eqref{eq:01-deriv-gamhat-fmsm}. 

\subsection{Proof for Lemma~\ref{lem:Dm-is-dual-to-Ptil}}
\label{ss:proof-lem:Dm-is-dual-to-Ptil}
Apparently, the function
\begin{tsaligned}
  \tilde{D}(\vA)
  :=
  -\frac{\lambda}{2}
  \lVert\tilde{\vw}(\vA)\rVert^{2}
  -
  \frac{1}{n}
  \sum_{i=1}^{n}
  \tilde{\Phi}^{*}(-\valph_{i}\,;\,y_{i}) 
\end{tsaligned}
is a Fenchel dual to $\tilde{P}$.
The first term of the right hand side in the
above equation can be rewritten as
\begin{tsaligned}
  \lVert\tilde{\vw}(\vA)\rVert^{2}
  =
  \lVert\vw(\vA)\rVert^{2}
  +
  \frac{\gamma_{\text{sm}}}{\lambda n}
  \lVert\vA\rVert^{2}_{\text{F}} 
\end{tsaligned}
which allows us to rearrange $\tilde{D}(\vA)$ as
\begin{multline}
  \tilde{D}(\vA)
  =
  -\frac{\lambda}{2}
  \lVert\vw(\vA)\rVert^{2}
  -
  \frac{\gamma_{\text{sm}}}{2 n}
  \lVert\vA\rVert^{2}_{\text{F}}
  -
  \frac{1}{n}
  \sum_{i=1}^{n}
  \left(
  \tilde{\Phi}^{*}(-\valph_{i}\,;\,y_{i})
  +
  \frac{\gamma_{\text{sm}}}{2}
  \lVert\valph_{i}\rVert^{2}
  \right)
  \\
  =
  -\frac{\lambda}{2}
  \lVert\vw(\vA)\rVert^{2}
  -
  \frac{1}{n}
  \sum_{i=1}^{n}
  \Phi_{\text{m}}^{*}(-\valph_{i}\,;\,y_{i})
  =
  D_{\text{m}}(\vA). 
\end{multline}
Hence, $D_{\text{m}}$ has been proved to be
a Fenchel dual to $\tilde{P}$.


\end{document}